\documentclass[letterpaper, 10 pt, conference]{ieeeconf}  

\IEEEoverridecommandlockouts                              

\usepackage{mathptmx} 
\usepackage{amsmath} 
\usepackage{amssymb}  
\usepackage{array}
\usepackage{xcolor}
\usepackage{graphicx}
\usepackage{subfigure}
\usepackage{xspace}
\usepackage{hyperref}

\title{\LARGE \bf
Handling Object Symmetries in CNN-based Pose Estimation*
}

\author{Jesse Richter-Klug$^{1}$ and Udo Frese$^{1}$
\thanks{*The research reported in this paper has been (partially) supported by the German Research Foundation DFG, as part of Collaborative Research Center 1320 EASE - Everyday Activity Science and Engineering, University of Bremen (http://www.ease-crc.org/, subproject R02).}
\thanks{$^{1}$Faculty of Mathematics and Computer Science,
        University of Bremen, 283569 Bremen, Germany
        {\tt\small \{jesse,ufrese\}@uni-bremen.de}}%
}

\def\eg{\emph{e.g}.\@ }
\def\ie{\emph{i.e}.\@ }

\def\etal{\emph{et al}.}
\newcommand{\figref}[2][]{(Fig.\@~\ref{#2}#1)}
\newcommand{\inlinefigref}[2][]{\xspace{}Fig.\@~\ref{#2}#1}
\newcommand{\secref}[1]{(Sec.\@~\ref{#1})}
\newcommand{\floor}[1]{\ensuremath{\lfloor{#1}\rfloor}}
\newcommand{\mvs}[1]{\left(\begin{smallmatrix}#1\end{smallmatrix}\right)}
\newcommand{\Rot}{\operatorname{Rot}}
\newcommand{\sfig}[1]{{\footnotesize{}#1}}

\newcommand{\cart}{\operatorname{cart}}
\newcommand{\pol}{\operatorname{pol}}
\newcommand{\cyl}{\operatorname{cyl}}

\begin{document}

\maketitle
\thispagestyle{empty}
\pagestyle{empty}

\begin{abstract}

        In this paper,  we investigate the problems that Convolutional Neural Networks (CNN)-based pose
        estimators have with symmetric objects. We considered the value of the CNN's output representation when continuously rotating the object and found that it has to form a closed loop after each step of symmetry. Otherwise, the CNN (which is itself a continuous function) has to replicate an uncontinuous function. On a 1-DOF toy example we show that commonly used
        representations do not fulfill this demand and analyze the problems
        caused thereby. In particular, we find that the popular 
        min-over-symmetries approach for creating a symmetry-aware loss tends not to work well with gradient-based 
        optimization, \ie deep learning.
    
        We propose a representation called ``closed symmetry loop'' (csl) from these 
        insights, where the angle of relevant vectors is multiplied by the
        symmetry order and then
        generalize it to 6-DOF. The representation extends our
        algorithm from \cite{richter2019towards} including a method
        to disambiguate symmetric equivalents during the final pose
        estimation.
        The algorithm handles continuous rotational symmetry (\eg a bottle) and discrete
        rotational symmetry (\eg a 4-fold symmetric box). It is evaluated on the T-LESS dataset, where it reaches state-of-the-art   for  unrefining  RGB-based  methods.
\end{abstract}


\section{Introduction}
\label{sec:intro}

Manipulating rigid objects at unknown poses has many applications,
from industry to household robotics. In the classical sense-plan-act
cycle, perception has to obtain the object poses, \eg from
a mono, stereo or depth camera image. This 6-DOF object pose problem is
well-studied in the ''vision for robotics`` field \cite{hodan2020bop},
nowadays successfully using deep learning with convolutional neural networks (CNNs).

\subsection{Challenges of Symmetric Objects}

A specific subproblem comes up, when the object is symmetric, either
in a continuous way (\eg a bottle) or in a discrete way (\eg a box or cube). Convolutional neuronal networks (CNNs) are continuous functions. As an object pose estimator, this function maps an image  to a likelihood of object existence and a set of Cartesian coordinates, which are describing the corresponding pose if it exists.
    A symmetrical object has multiple visually indistinguishable points. Consequential, there are multiple sets of Cartesian coordinates that are describing different but equally valid poses.

    The properties of this functions depend on the representation
for the points resp. pose output. In this work, we
show that for discrete symmetrical objects and commonly used
representations this leads to uncontinuous functions. This is a contradiction to the CNN's abilities. Therefore, the CNN may only learn an approximation. We investigate in a 1-DOF toy problem, 
what effect this has for different representations and find that the popular 
min-over-symmetries approach~\cite{park2019pix2pose, wang2019normalized, wang2019densefusion, xiang2017posecnn}
 for a symmetry-aware loss tends not to work well with gradient-based optimization, \ie deep 
learning.

    Conversely, we derive a representation for the CNN's output space (closed symmetric loop) where symmetrical equivalent poses are mapped to the same values and the resulting function is continuous. Hence, we removed the uncontinuous part out of the CNN allowing it to learn the true mapping instead of a mere approximation. This is paired with a reverse transformation that yields a valid pose afterwards. 
We derive this representation and transformations from the toy example study and
generalize it to full 6-DOF.

\begin{figure}
    \centering
    \includegraphics[height=3cm]{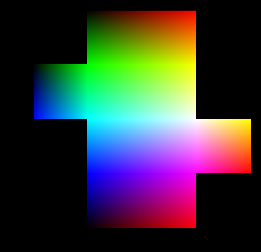}\quad
    \includegraphics[height=3cm]{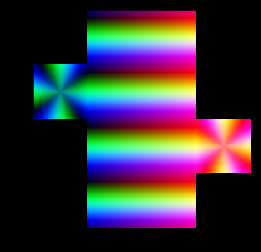}    
    \caption{Two representations of the surface points of a $4$-fold symmetric box as color coded $3D$-vectors (unwrapped): left plain
             object points, right the proposed \emph{closed symmetry loop} or \emph{star} representation. The right is continuous and
             respects symmetry, whereas the left does not.}
    \label{fig:repr}
\end{figure}

\subsection{This Work's Contribution and Structure}

\begin{itemize}
   \setlength{\itemsep}{0px}
   \item a mathematical analysis which properties a CNN
   output representation must have to represent the pose of a
   symmetric object in a continuous way \secref{s:mat_moti};
   \item an investigation with a 1-DOF toy problem that demonstrates
   the effect the continuity problem has for common representations
   and motivates a solution \secref{s:toyexample};
   \item an algorithm for 6-DOF pose estimation based on this idea,
   extending \cite{richter2019towards} to symmetric objects \secref{sec:approach} and
   \item an evaluation on the T-LESS benchmark dataset showing
   competitive results (RGB 46.8, RGBD 58 AR) \secref{sec:exp_6dof}.
\end{itemize}
Finally, Section~\ref{sec:rel_work} relates the observations to prior work and Section~\ref{sec:concl} concludes. 
The source code of this work is available\footnote[2]{https://github.com/jesserichterklug/closed-symmetry-loop}.

\section{Mathematical Motivation}
\label{s:mat_moti}

This section motivates the approach to define the output of the CNN as a specialized
representation that reflects the symmetry of the underlying object and derives what structure
this representation needs to have. Consider an object with $n$-fold, \ie $\theta=\frac{2\pi}{n}$,
rotational symmetry around the $Z$-axis. Let $\Rot_z(\alpha)$ be rotation around $Z$ and $f$
be a ``render'' function that maps for a fixed object and scene,
 a pose to an image of
the object in that pose. Since the object is symmetric 
\begin{gather}
    f(T) = f\left(T\Rot_Z\left(i\theta\right)\right)\quad\forall T\in SE(3), i\in\mathbb{Z}
\end{gather}
Note that $f$ is continuous, as small changes in pose lead to small changes in the image.
Now let $g$ be the function learned by the CNN, mapping from an image to some 
representation of the pose by real numbers $\in\mathbb{R}^m$. 
Examples from the literature are a matrix, a quaternion, a heatmap of bounding-box corners \cite{park2019pix2pose}, object-coordinates per pixel \cite{pavlasek2020parts} or any other suitable representation. Now being a CNN,
$g$ is continuous and $f$ is continuous as well, so for a given $T\in SE(3)$,
\begin{gather}
    h: \left[0\ldots \theta\right] \rightarrow \mathbb{R}^m, \quad
    \alpha \mapsto g(f(T \Rot_Z(\alpha)))
\label{eq:scc}
\end{gather}
is a continuous function. It is also injective except for $0$ and $\theta$ because
all poses in between are not equivalent even with symmetry. So $h$, \ie the pose representation for continuously rotating by one step of symmetry, is a simple
closed curve. This is not possible for any above mentioned representation, where rotating
by $2\pi$ is a simple closed curve but by $\theta$ is not. Note that this is true,
regardless whether the pose representation is ``interpreted modulo $\theta$'' later,
because CNNs cannot represent functions that are continuous in some modulo topology but
not in the usual $\mathbb{R}^m$ topology.

Of course a CNN can also learn to approximate an uncontinuous function. Probably it will be
 steep (but still continuous) at a gap of the training data, since that does not affect the training loss. So we can conclude
that by choosing a pose representation that does not reflect the objects $n$-fold symmetry,
we force the network to approximate an uncontinuous function and give rise to generalization
problems.


\section{1-DOF Toy Problem Investigation}
\label{s:toyexample}

\begin{figure}
    \newlength\myw
    \setlength{\myw}{0.42\columnwidth}
    \center
    \renewcommand{\arraystretch}{0.5}
    \begin{tabular}{cc}
    \includegraphics[width=\myw]{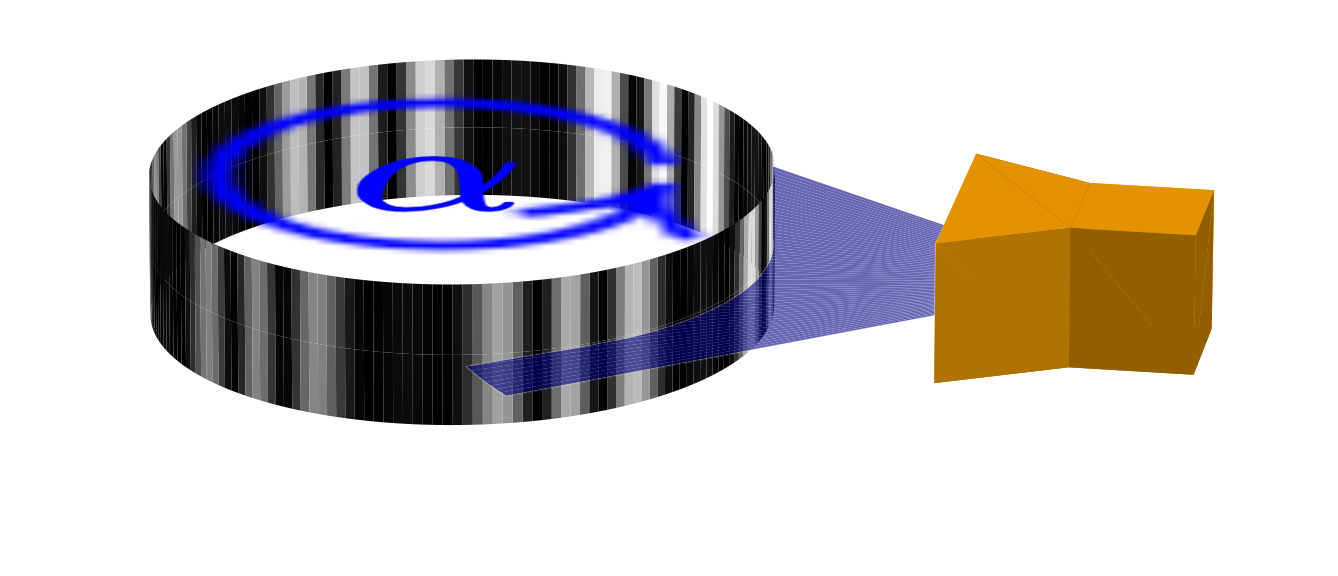} 
    & 
    \includegraphics[width=\myw]{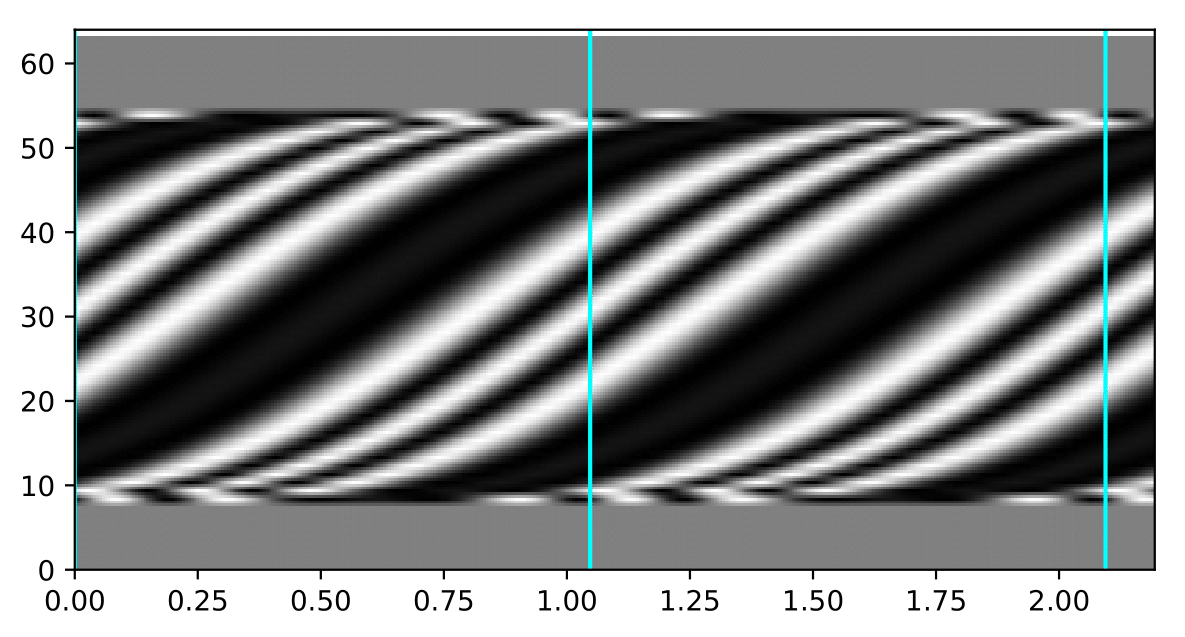} \\
    \sfig{a) disc and camera\hspace{5mm}}&
    \sfig{b) 1-D images over $\alpha$}\\
    \\ \hline
   \includegraphics[width=\myw]{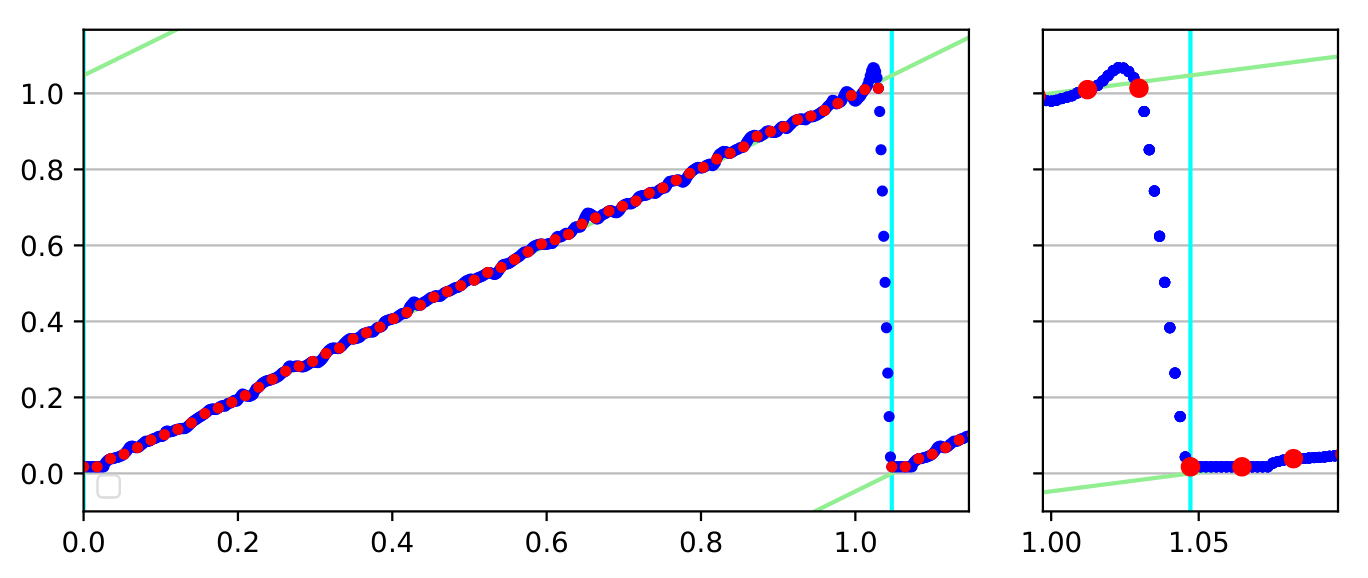} &
    \includegraphics[width=\myw]{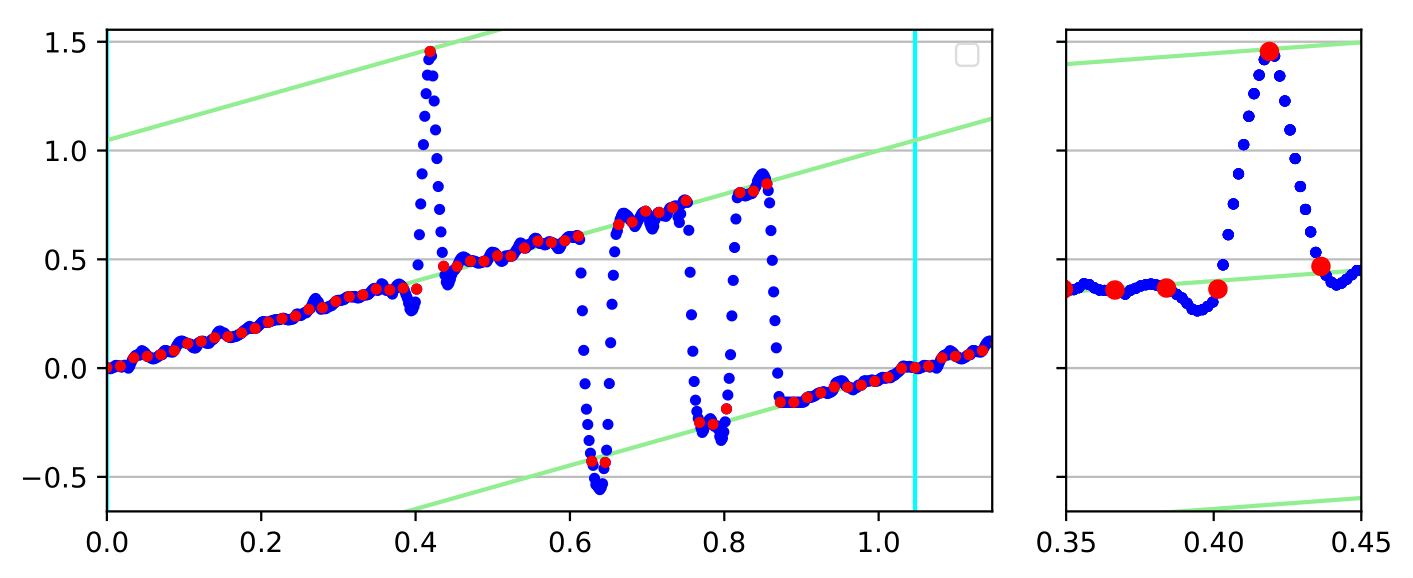}   \\
     \sfig{c) norm. angle / ae}& 
     \sfig{d) angle / mos-ae}\\
\\
   \includegraphics[width=\myw]{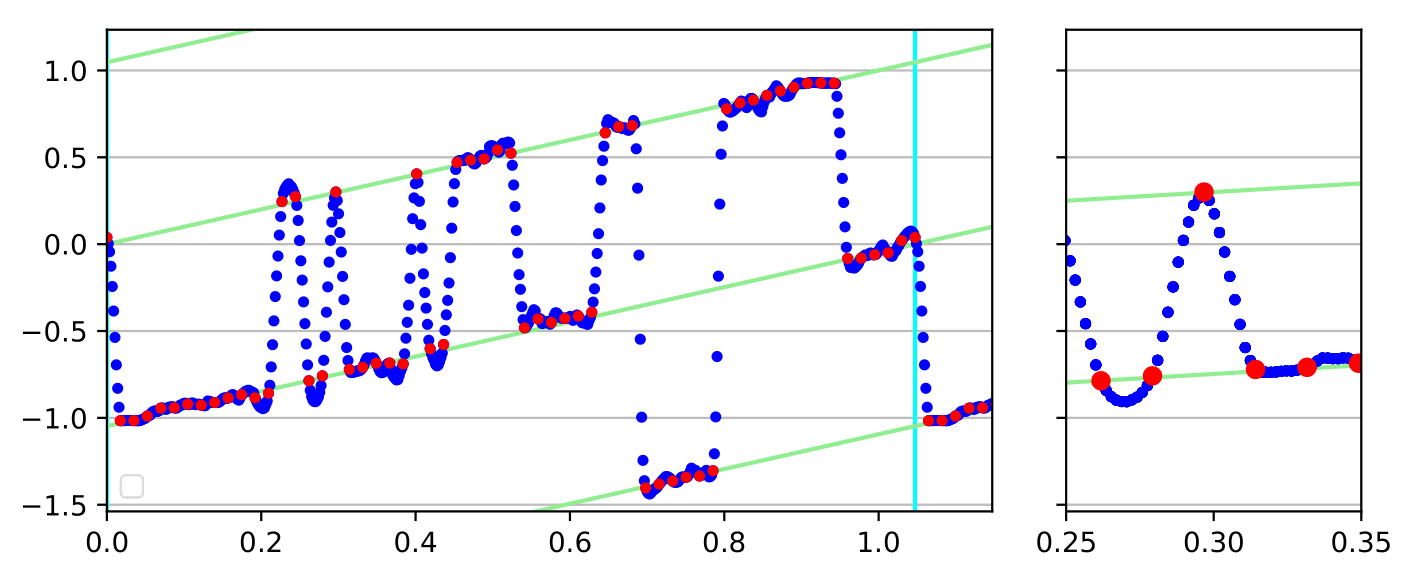} &
    \includegraphics[width=\myw]{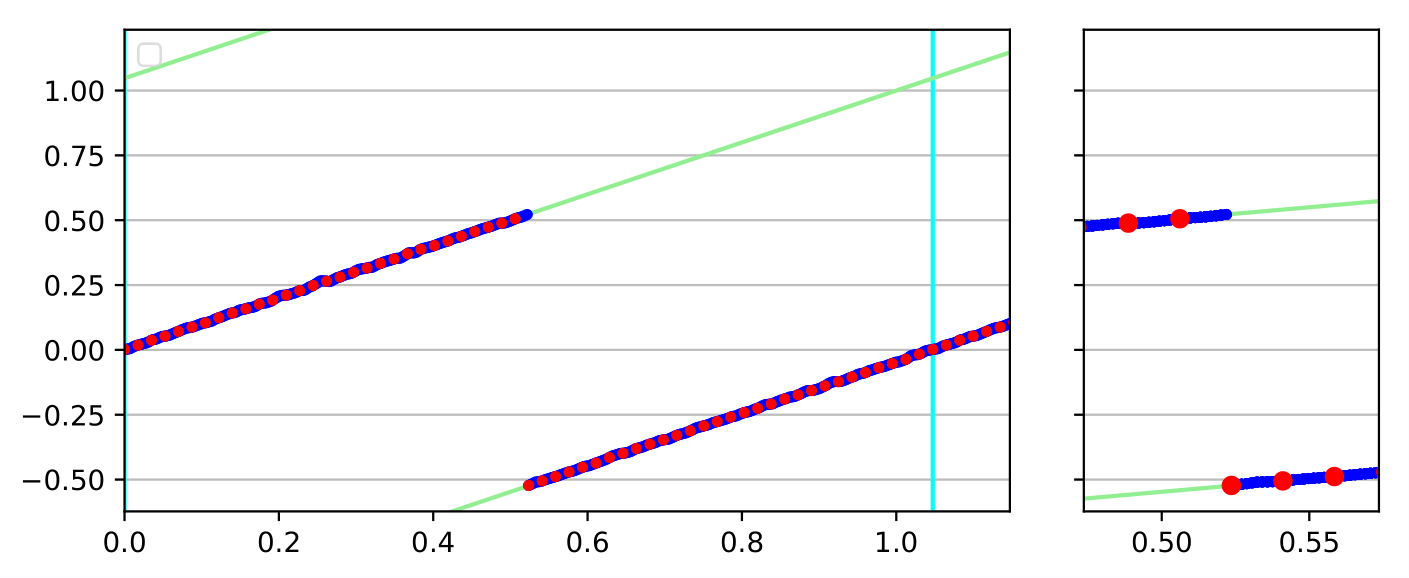}   \\
   \sfig{e) vector / mos-ae}& 
   \sfig{f) csl-vector / ae}\\
\\
    \hline
    \includegraphics[width=\myw]{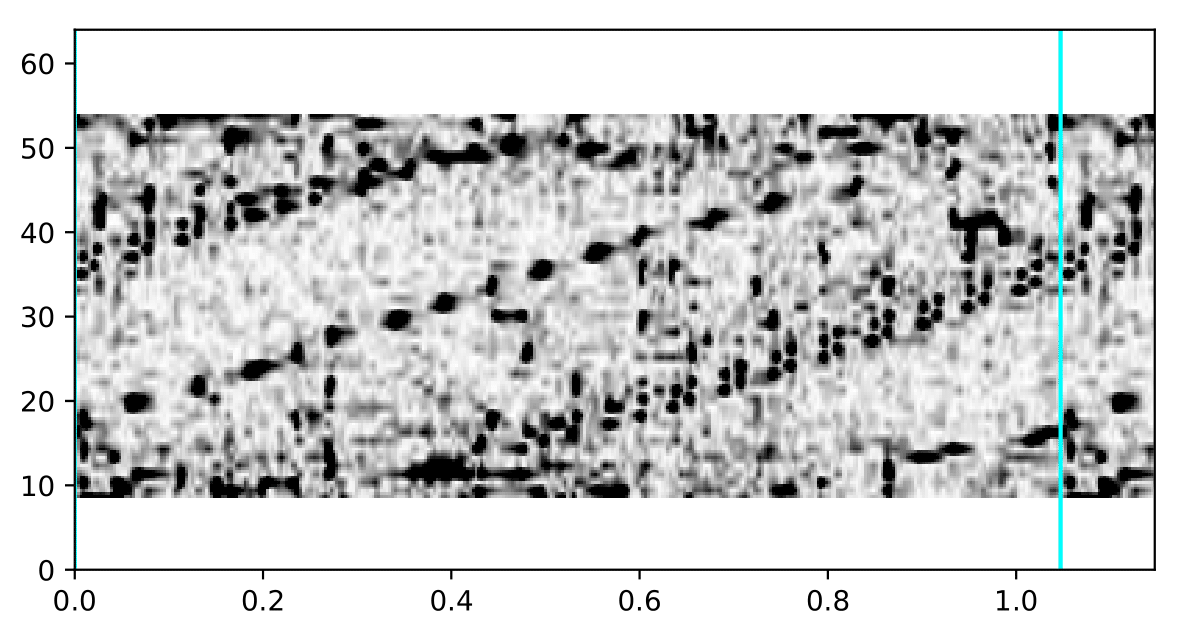} &
    \includegraphics[width=\myw]{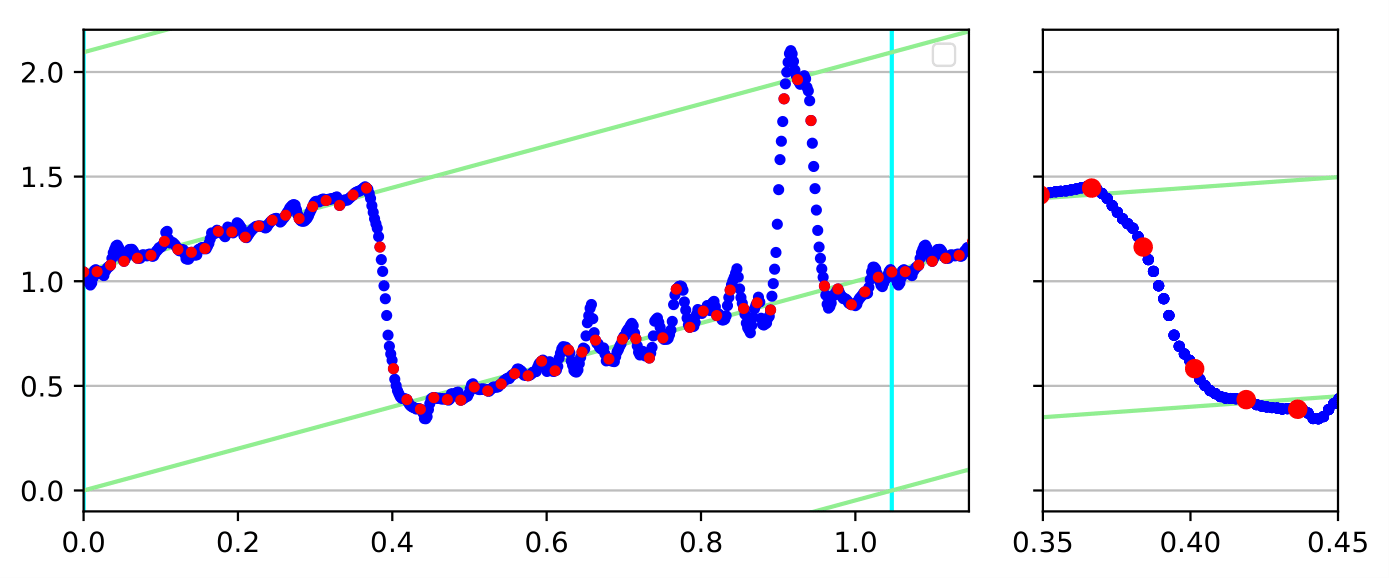} \\
    \sfig{g) $p^O$-img / px-mos-ae}& 
    \sfig{h) center pixel of g)} \\
   \includegraphics[width=\myw]{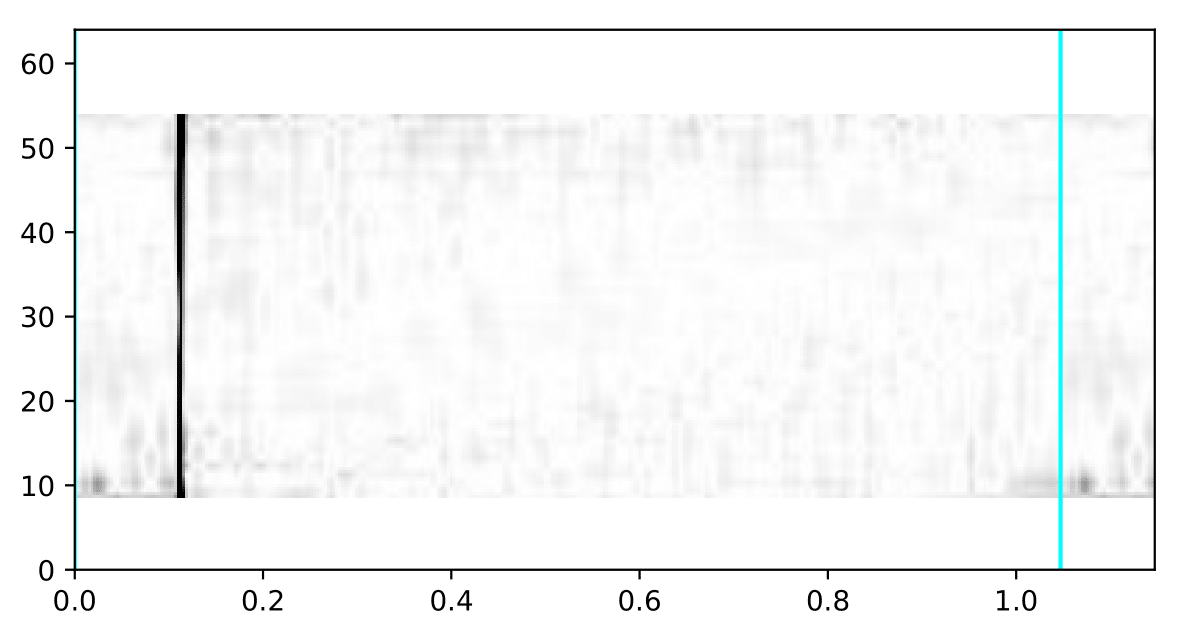} &
    \includegraphics[width=\myw]{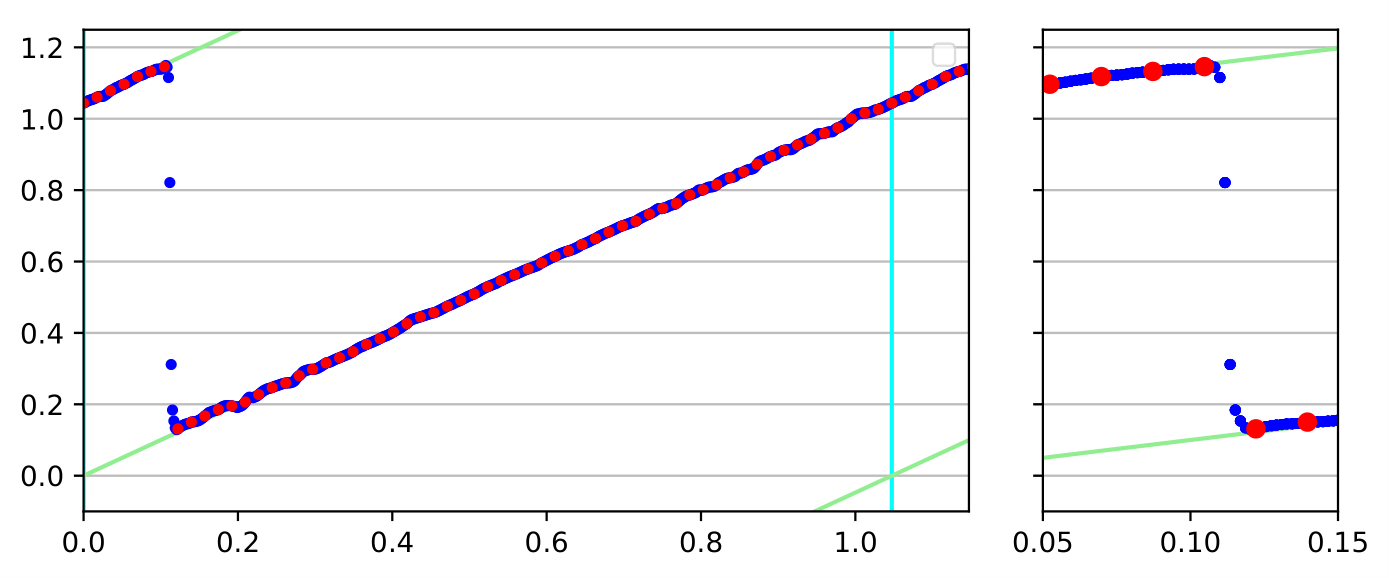} \\
    \sfig{i) $p^O$-img / img-mos-ae}& 
    \sfig{j) center pixel of i)}\\
    \includegraphics[width=\myw]{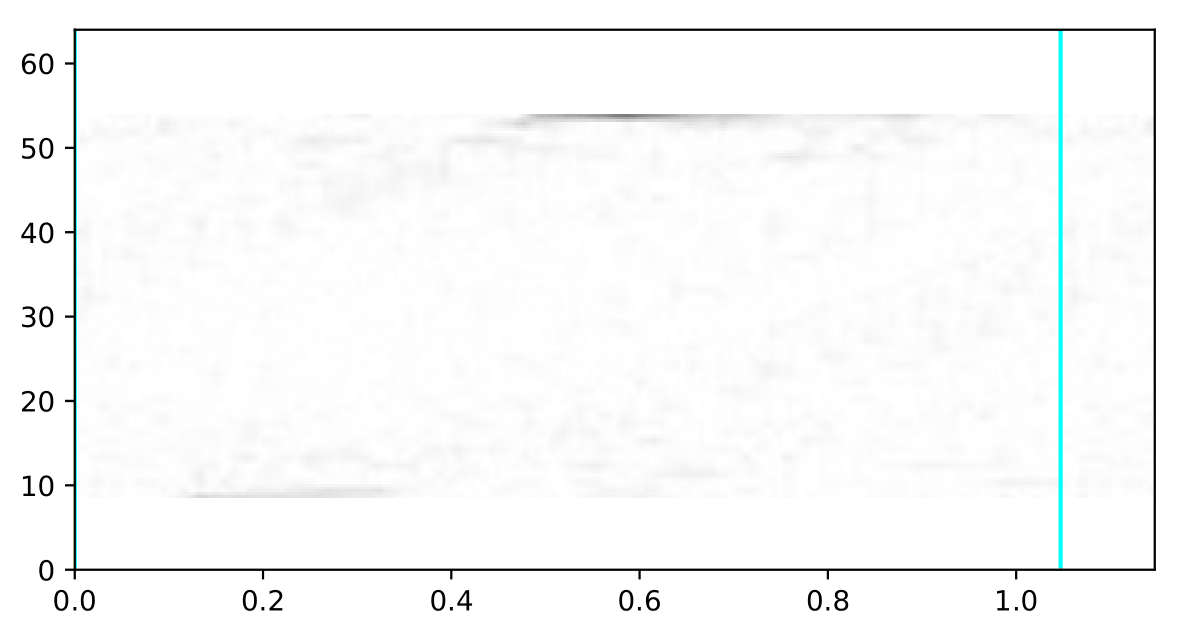} &
    \includegraphics[width=\myw]{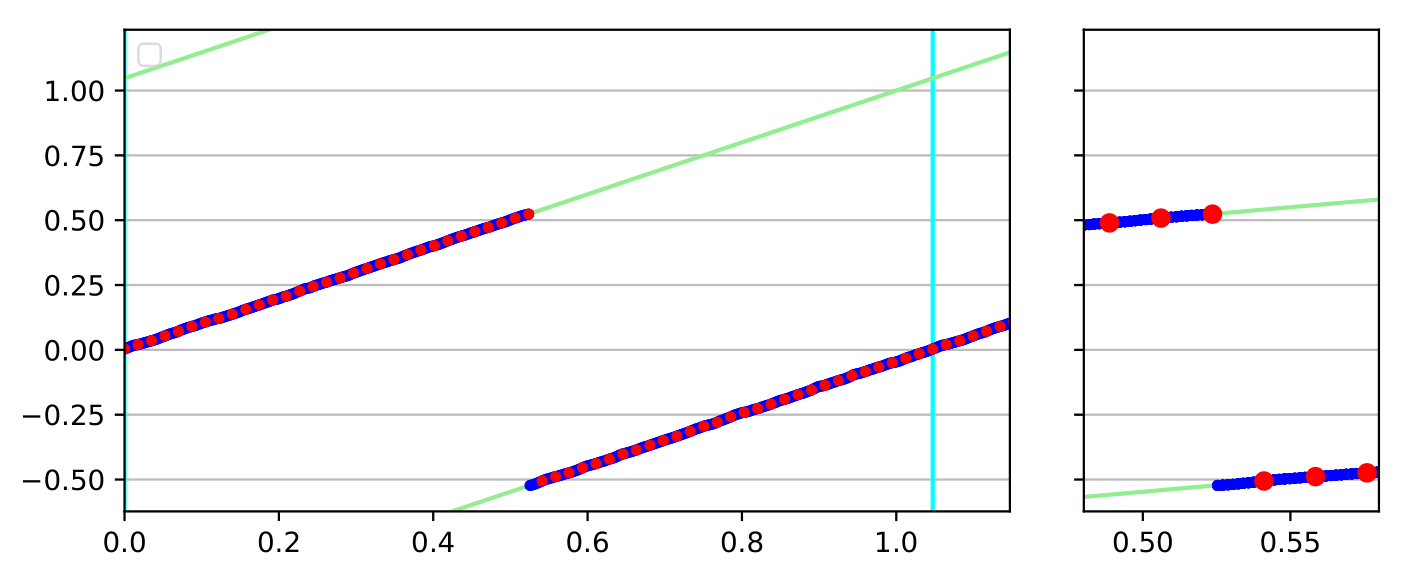} \\
    \sfig{k) $p^O$-img / ae}& 
    \sfig{l) center pixel of k)}
    \end{tabular}
    \caption{Comparing different output representations for the angle of a rotating disc with 6-fold symmetric texture.
    In all plots the ground truth angle is shown on the x-axis and the cyan vertical lines indicate periodicity, the ground truth and its symmetric equivalents 
    are shown in green, the CNN prediction converted
    to an angle in blue and the prediction on training data is highlighted in red. g/i/k show the error of the represented object
    points (black=large).
See Sec.\@~\ref{s:toyexample} for details.
    }
    \label{fig:toyexample}
\end{figure}

We will now analyze a toy problem that is simple enough,
so we can plot the CNN's behavior on the whole input data, 
but still exhibit the above mentioned phenomenon: A rotating disc with textured perimeter
is viewed from the side by a line-camera 
\figref[a]{fig:toyexample}.
The disc's texture has an $n=6$-fold symmetry, \ie the angle of symmetry is $\theta=2\pi/n=\pi/3\approx1.05$ (cyan lines in 
\inlinefigref{fig:toyexample}). From the obtained 1D-image
\figref[b]{fig:toyexample},
a CNN shall estimate the rotation angle $\alpha$ of the disc as $\hat{\alpha}$. We
are interested in how well the CNN can learn this task for different
representations of the angle as output and different corresponding
losses.

As the focus is on the output representation and the problem is rather
simple, we use a canonical encoder-head architecture, details can be seen in the implementation.
Our training dataset has images at $\pi/180$ spaced
angles, the test set at $\pi/900$ spaced. We trained every CNN 11 times and
report on the network with the median loss.

\subsection{Outputs representing an angle}

We initially consider the task to estimate $\alpha$ the disc
orientation in some representation.

\medskip 

The first output representation is \textbf{normalized angle}  $\in[0\ldots\theta[$ with an absolute error loss:
\begin{equation}
    y^\text{norm. angle} = \alpha - \theta\floor{\tfrac{\alpha}{\theta}}, \quad \mathcal{L}^\text{ae}(y, \hat{y}) =  |y-\hat{y}|,
    \label{eq:normangle}
\end{equation}

where $y$, $\hat{y}$ and $\mathcal{L}$ are groundtruth output, predicted output and loss of a sample, which has groundtruth disc angle $\alpha$. The representation forms no closed loop, so the CNN has to 
approximate the discontinuity at $\theta$ by a steep transition.
It does so \figref[c]{fig:toyexample} by placing the transition between two training
samples, so it is invisible in the loss, but creates a small region
of large (up to $\theta/2$) generalization error.

\medskip 

The second idea is to use the \textbf{angle}, but interpret it
``modulo $\theta$'' by viewing it as the set of all symmetric
equivalents. Canonically, the distance to a set is defined as
minimum distance over its elements. This leads to the minimum-over-symmetries absolute error (mos-ae) loss:
\begin{equation}
    y^\text{angle} = \alpha, \quad 
    \mathcal{L}^\text{mos-ae}(y, \hat{y}) = \min_{k\in\mathbb{Z}} | y-\hat{y}+k\theta|
    \label{eq:angle}
\end{equation}

This appears like an elegant solution. However, it does not form a
closed loop as the output at $0$ and $\theta$ is not equal but
only equivalent. So it also requires the CNN to learn a 
discontinuity creating a transition. The experimental result is even worse, making many apparently unnecessary transitions on the
way~\figref[d]{fig:toyexample}. Presumably, these appear when in
early learning stages different symmetric equivalents of the 
groundtruth  are closest and the loss pulls the CNN towards these.
Later, the solution can not move from one equivalent to another,
as they are separated by a barrier of large loss. This observation sheds doubt on the effectiveness of the
minimum-over-symmetries approach. 

\medskip 

The third idea replaces the angle by a unit \textbf{vector} to
eliminate the $2\pi$-wraparound:
\begin{align}
    y^\text{vector} &= \cart\mvs{\alpha\\1}, \; 
    \mathcal{L}^\text{mos-ae}(y, \hat{y}) = \min_{k\in\mathbb{Z}}
     |\Rot(k\theta)y-\hat{y}|,         
     \label{eq:vector}\\ 
     &\text{with }
     \cart\mvs{\phi\\\rho} = \mvs{\cos\phi \rho\\\sin\phi \rho}\!,\;
     \Rot{\phi} = \mvs{\cos\phi& -\sin\phi \\ \sin\phi & \cos\phi}
\end{align}

It is still not a closed loop and $2\pi$ is actually
not the problem, because $\theta$ is. Correspondingly, this approach
performs not better than the previous \figref[e]{fig:toyexample}.

\medskip 
 
The proposed \textbf{closed symmetry loop (csl) vector} representation starts from the observation that 
the vector representation forms a closed loop over $[0\ldots2\pi]$.
Hence, we multiply the angle by $n$ before turning it into a vector. So $\theta$ becomes $2\pi$ and the
csl vector forms a closed loop over $[0\ldots\theta]$. The 
representation respects symmetry, mapping symmetric
equivalents to the same value:
\begin{equation}
    y^\text{csl vector} = \cart\mvs{n\alpha\\1}, \quad
    \mathcal{L}^\text{mos-ae}(y, \hat{y}) = |y-\hat{y}|
    \label{eq:cslv}
\end{equation}

With this representation the CNN learns a function without
transitions \figref[f]{fig:toyexample}. Note that the discontinuity
in the graph comes from converting the vector back to an angle for plotting and does not appear in the output itself.

\subsection{Outputs Representing an Object Point Image}

We now turn towards a more complex but related problem, which we
need in \cite{richter2019towards} and Sec.\@~\ref{sec:approach} later. Here,
the output is an image, where each pixel indicates the point of the
object seen in that pixel in object coordinates ($p^O$-image). So the CNN answers
the question ``What do you see here?'' and the final object pose is
obtained by a perspective n-point (PnP) problem from that. Different
reprensentations and corresponding losses for $p^O$ are possible,
which we will investigate here.
We therefor extend the CNN to a canonical encoder/decoder
with shortcuts architecture.

\medskip 

The first idea uses a \textbf{$p^O$-image}
representation where each pixel of the output is the 2D vector
of the seen point in object coordinates. Symmetry is again
handled by a min-over-symmetries loss. It takes the
average over all pixels of the minima (\textbf{pmos-mae}), thereby allowing each pixel to choose its own symmetric equivalent. 
\begin{equation}
    y^\text{$p^O$-img}_{i} \!\!=\!\! p^O_{i}, \; \mathcal{L}^\text{pmos}_\text{-mae}(y, \hat{y}) = 
       \tfrac{1}{m} \sum_i \min_{k\in\mathbb{Z}} |\Rot(k\theta)y_{i}-\hat{y}_{i}|, 
    \label{eq:opipmos} 
\end{equation}

where $p^O_{i}$ is the true point of the disc visible 
at pixel $i$ and $m$ is the number of pixels.
\inlinefigref[g/h]{fig:toyexample} show the result with a large error and many unnecessary transitions. 

\begin{table}
\center
\vspace{5px}
\begin{tabular}{llcc}
    \textbf{output repres.} & \textbf{loss} & \textbf{pixel error} & \textbf{angle error} \\
    normalized angle & ae \eqref{eq:normangle} & & 0.0099\\
    angle & mos-ae \eqref{eq:angle} & & 0.0378\\
    vector & mos-ae \eqref{eq:vector} & & 0.0660\\
    csl vector & ae \eqref{eq:cslv} & & \textbf{0.0020} \\ \hline
    $p^O$ image & pmos-mae \eqref{eq:opipmos} & 0.0703 & 0.0092\\
    $p^O$ image & imos-mae \eqref{eq:opiimos} & 0.0074 & 0.0045 \\
    csl image & mae \eqref{eq:opicsl} & \textbf{0.0029} & \textbf{0.0005} \\
\end{tabular}
\caption{Error of CNNs with different output representations. Abbr.: (p/i)mos - (pixel/image) min over symmetries,
csl - closed symmetry loop (proposed), $p^O$: object point }
\label{tab:results}
\end{table}

\medskip 

The second idea also uses a \textbf{$p^O$-image} but takes
the min of the averages, \ie per image (\textbf{imos-mae}). This forces consistency, \ie all
pixels choose the same  equivalent.
\begin{align}
    y^\text{$p^O$-img}_{i} = p^O_{i}, \; \mathcal{L}^\text{imos}_\text{-mae}(y, \hat{y}) = 
       \tfrac{1}{m} \min_{k\in\mathbb{Z}} \sum_i |\Rot(k\theta)y_{i}-\hat{y}_{i}| \!\!\!
    \label{eq:opiimos}
\end{align}
\inlinefigref[i/j]{fig:toyexample} show that imos-mae
is much better than pmos-mae. This is surprising,
because the optimal $k$ from \eqref{eq:opiimos} is also
a valid choice for all $i$ in \eqref{eq:opipmos}.
Thus $\mathcal{L}^\text{pmos-mae}(y, \hat{y}) \le 
\mathcal{L}^\text{imos-mae}(y, \hat{y})$. 
However, as with mos-ae,
pmos-ae attracts the CNN early to different symmetric 
equivalents, creating unnecessary transitions. By forcing
consistency in one image, it also supports
consistency over angles,
because images at similar angles mainly differ by a translation
for which a CNN is invariant. Still, it forms no closed loop, the CNN has to learn one discontinuity and
there is one transistion because of that. \inlinefigref[i]{fig:toyexample} shows that all pixels perform this transition at
the same angle, to maintain consistency in the images.

Finally, the proposed \textbf{csl image} representation for the $p^O$ also called $p^{O*}$ forms a closed loop when continuously rotating by $\theta$
and can use a simple \textbf{mae} loss.
It takes the $p^O$ vector in every pixel and multiplies its angle
by $n$. As with the csl vector representation, a rotation by $\theta$ is mapped to a
rotation by $2\pi$, which is a closed loop.
\begin{align}
    \notag
    y^\text{csl img}_{i} \!&=\! p^{O*}_{i} \!=\!
    \cart\bigl(\mvs{n&\\&1} \pol\mvs{p^O_i} \bigr),\;
    \pol\mvs{x\\y} = \mvs{\operatorname{atan2}(y,x)\\\sqrt{x^2+y^2}}, \\
    &\mathcal{L}^\text{ae}(y, \hat{y}) = 
       \tfrac{1}{m} \sum_i |y_{i}-\hat{y}_{i}| 
    \label{eq:opicsl}
\end{align}
\inlinefigref[k/l]{fig:toyexample} show that there is no transition,
the visible discontinuity comes again from plotting the result as
an angle.

\subsection{Discussion}

Table \ref{tab:results} compares all representations quantitatively.
If the representation is an image, both the average per pixel error
and the error of the final angle is given. This is obtained by simply
comparing to a precomputed list of object point images with interpolation. In the 3D scenario,
later, this is a more complex PnP problem. 

We conclude with three insights: First, minimum-over-symmetries losses, while 
mathematically elegant, tend to not work well with gradient-based
optimization of a CNN. Second, letting the CNN output an object point
image from which the pose is geometrically computed is more precise
than letting the CNN directly output the pose.
Third, by multiplying the angle of
a vector with the order of the symmetry, we can define the star
representation that forms a closed loop and makes the function
to be learned continuous and that achieved the lowest error in this
study.

\section{Approach (6-DOF)}
\label{sec:approach}

Following the above considerations, we modified our previous representation \cite{richter2019towards} 
in a symmetry-specific way, such that rotating by one step of symmetry, \ie $\theta=\frac{2\pi}{n}$, is a simple closed curve in the representation.

\begin{figure}
    \center
    \vspace{2px}
    \includegraphics[width=\columnwidth]{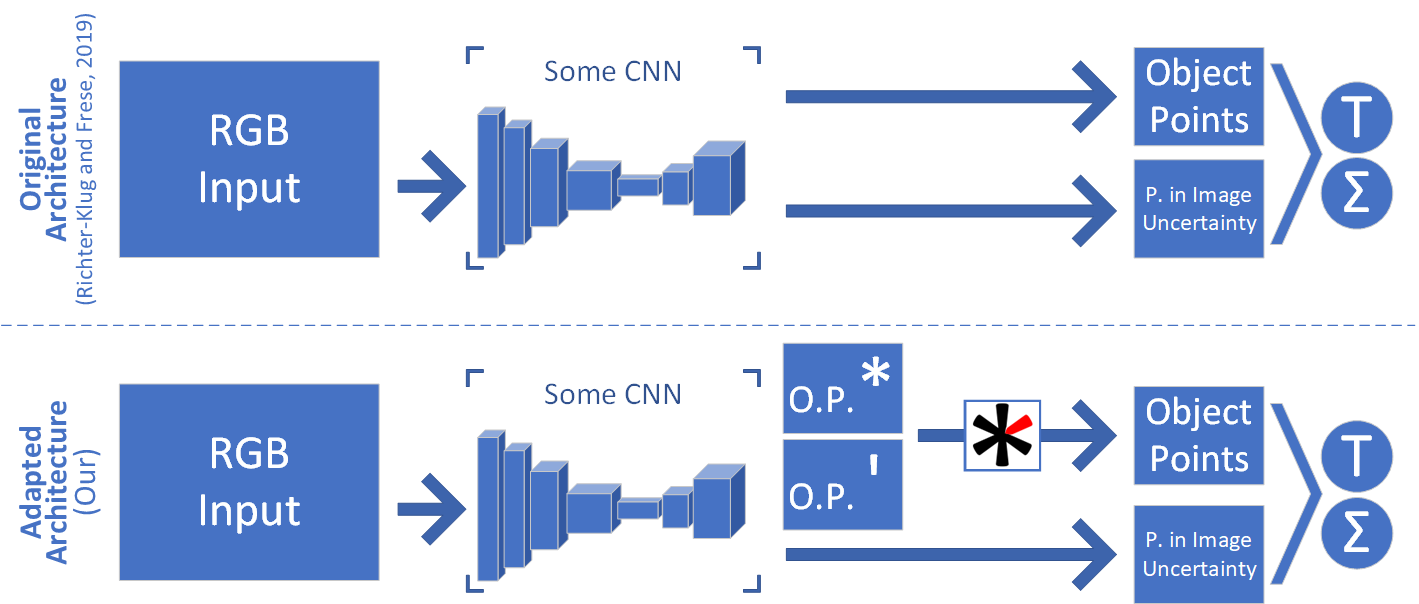}
    \caption{Network architecture extension overview adapted from \cite{richter2019towards}. Originally (top), an RGB image is fed into a CNN, which outputs the seen object point (per pixel) as well as an estimate of their in-image uncertainties. This information is then combined by PnP with all the pixels that belong to the same object to estimate its pose ($T$) and 6d uncertainty ($\Sigma$).
    In this paper (bottom), we adapt this architecture with a symmetry-aware but ambiguous object point representation (star), which is aided by the dash representation, both predicted by a CNN. They are then
    combined to regain the object points, followed by the unchanged PnP stage.}
    \label{fig:expantion_overview}
\end{figure}

In the originally proposed architecture, the CNN predicted object points densly. These were regressed by PnP for getting a pose estimate. In addition, the CNN predicted in-image uncertainty for each found object point. Therefore, the PnP could also provide a 6d uncertainty estimate (Fig. \ref{fig:expantion_overview}-top).

\begin{figure}
    \center
    \vspace{5px}
    \includegraphics[width=\columnwidth]{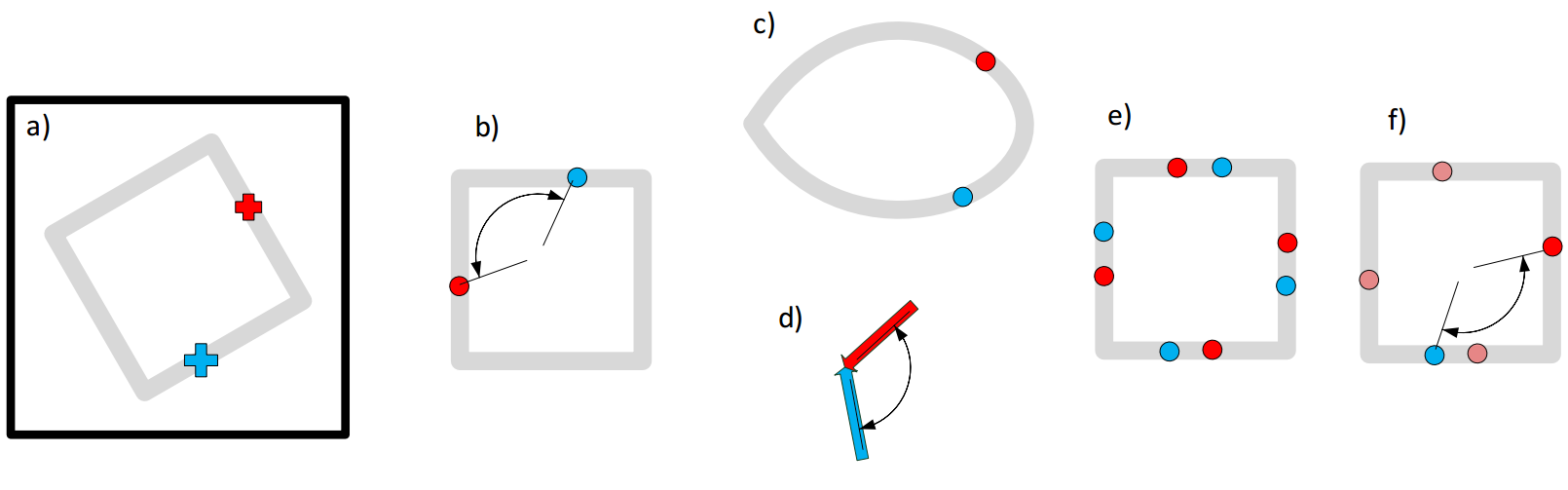}
    \caption{Steps of the forth and back transformation with two 
        points marked as examples. All quantities are
        actually 3D vectors, here we show X and Y for clarity,
        Z is the axis of symmetry. \textbf{a)}
        the image perceived by a camera looking on a box from above. 
        \textbf{b)}
        object points $p^O$ as used in \cite{richter2019towards}, 
        \textbf{c)} $p^{O*}$
        information predicted by the CNN,  
        \textbf{d)} $p^{O'}$ information
        also predicted by the CNN, 
        \textbf{e)} $P^O$
        equivalence classes obtained from $p^{O*}$,
        \textbf{f)} consistent 
        disambiguation of the $P^O$ using $p^{O'}$ to regain
        $p^o$. (blue:arbitrarily chosen reference $p_r$, red: best fitting
        point from equivalence class $P^O$)
    }
    \label{fig:steps}
\end{figure}

\begin{figure}
    \setlength{\myw}{0.27\columnwidth}
    \center
    \begin{tabular}{ccc}
    \includegraphics[width=\myw]{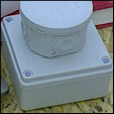} 
    & 
    \includegraphics[width=\myw]{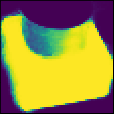}
    & 
    \includegraphics[width=\myw]{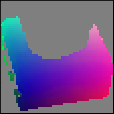} \\
    \sfig{a) RGB input}&
    \sfig{b) segmentation}&
    \sfig{c) true object points}\\
    \\ 
    \includegraphics[width=\myw]{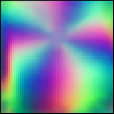} 
    & 
    \includegraphics[width=\myw]{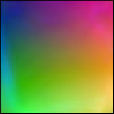}
    & 
    \includegraphics[width=\myw]{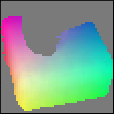} \\
    \sfig{d) star output}&
    \sfig{e) dash output}&
    \sfig{f) new object points}
    \end{tabular}
    \caption{Example input image a) with object segmentation b) and the unknown true object points c). The proposed reverse operation uses our outputs d) and e) to genereate the object points f). These are then used to estimate the object's pose. Note that f) is not equal to c) but it could have been. In this specific case, it is instead offsetted by two steps of symmetry.
    }
    \label{fig:ex_output_images}
\end{figure}

To make this architecture symmetry-aware, we change the CNN's object point output to a symmetry-aware one, the so-called star representation (Sect.\@\ref{s:po_star}), and regain valid object points before the PnP stage (but outside of the CNN) by reversing the representation's modification (Sect.\@\ref{s:reverse_op}). A second CNN output, the so-called dash representation (Sect.\@\ref{s:po_dash}) helps by untangling the object point ambiguities caused by the symmetry (Fig. \ref{fig:expantion_overview}-bottom,
Fig. \ref{fig:steps}, Fig. \ref{fig:ex_output_images}).  

\subsection{The star representation of object points}
\label{s:po_star}

The representation is a modification of the object points such that rotating by one step of symmetry, \ie $\frac{2\pi}{n}$, is a simple closed curve in the representation (csl-image). In it, all object points, that appear the same (based on the defined symmetry), are mapped on the same value and no possible rotation will result in an uncontinuous change. Therefore, the representation becomes symmetry aware, but also ambiguous. 

To gain the star representation of the object points, these are first transformed in cylindrical coordinate space, where the cylindric axis is aligned with the symmetry axis. Here the angle value is multiplied by $n$ (the fold of symmetry). Afterwards the points are transformed back to Cartesian vector space \figref[c]{fig:steps}.
\begin{align}
    p^{O*}_{ij} &= \cart\Bigl( \mvs{n&&\\&1&\\&&1} \cyl(p^O_{ij}) \Bigr), 
        \label{eqn:postar}
\\
    \text{with }
    \cart\mvs{\rho\\\psi\\z} &= \mvs{\rho \cos \psi \\ \rho \sin \psi \\ z}, \quad
    \cyl\mvs{x\\y\\z} = \mvs{\operatorname{atan2}(y,x) \\ \sqrt{x^2+y^2} \\ z}
\end{align}

For clarity, this assumes, w.l.o.g. Z as symmetry axis. 

Note that the CNN is trained to output $p^{O*}$ so the computation
in \eqref{eqn:postar} is not executed when using the algorithm but
when preparing the ground truth output for training.

Let's have a closer look at the folds of symmetry extremes: On the lower end, one finds non-symmetrical objects ($n=1$); In this case the star representation is identical to the origin object points which is the expected outcome. On the other end, we find objects with infinity-fold symmetries, \eg bottles. Here an infinitely small step of rotation closes one step of symmetry. Since the multiplication with infinity is unhandy, in this case, we multiply the angle values with zero. Therefore, all points have the same angle around the rotation axis as they all are equivalent under
symmetry.


\subsection{The dash representation of object points}
\label{s:po_dash}

The ambiguity of the star representation causes ignorance whether two points, whose values are close, also lie close on the object or \eg on opposing ends. But, this information is needed to regain an object point that is consistent with all points in view (cf. \ref{s:reverse_op}). We argue that this information can be seen inside an image despite or rather independently of any possible symmetries and therefore is extractable. 

As such information we use the pixelwise object points rotated into the camera. This is minus the vector from the object point to the object's origin relative to the camera. We argue that this vector is observable in the image and hence can be predicted by
a CNN. Note, this information is innately symmetrical invariant and (since we only rotated the object points) all angles between any object points are preserved, but no information regarding the object's rotation itself \figref[d]{fig:steps}.  

The selected information can not be learned as is, since orientation is not a translation invariant function of the image (cf. \cite[Fig. 2]{richter2019towards}). Thus, depending on the
pixel position in the image, we rotate the vector, such that the
CNN can treat it as if in the image center. Formally,
%
%
\begin{align}
    p^{O'}_{ij} &= R_{ray}^{-1}(i,j) \; R^C_O \; p^O_{ij}, \\
    R_{ray}(i,j) &= 
    {\;}^{\operatorname{angle}}_{\operatorname{axis}}\biggl(\sphericalangle 
         \Bigl(\mvs{0\\0\\1}, \operatorname{ray}(i,j)\Bigr)
         , \mvs{0\\0\\1} \times \operatorname{ray}(i,j)\biggr)
\end{align}

$R_{ray}(i,j)$ is a matrix rotating the Z-axis onto the viewing ray of pixel $(i,j)$. The viewing rays are defined by the camera calibration.

Note that before this representation's usage (\ie \ref{s:reverse_op}) the rotational offset must be reversed.

\subsection{The reverse operation}
\label{s:reverse_op}

The purpose of the reverse operation is to gain an image of object points that together define a pose in the PnP stage that is right up to the object's symmetry. Each point in the star representation defines an equivalence class of object points \figref[e]{fig:steps} that can be extracted by reversing \eqref{eqn:postar} as
\begin{align}
    P^O_{ij} \!=\! \left\{\!\cart\biggl(\!\!
    \mvs{\tfrac{1}{n}&&\\&1&\\&&&1}\!\!
    \cyl\Bigl(
         p^{O*}_{ij}
    \Bigr) \! +\! \mvs{k\theta\\0\\0}
    \biggr) \bigg| k \in [0\ldots{}n[\right\}\!.\!\!\!
    \label{eqn:equival}
\end{align}

Although each point of the equivalence class would be per se valid, only a consistent choice over all recognized points of an object will lead to a correct pose prediction. Two points are chosen consistently if their offset equals their true offset, \eg if two points oppose each other opposing object points must be chosen, too. To determine the offset between two object points, the dash representation was introduced. In it, the angle between two vectors is the same as between their corresponding object points, if selected consistently (cf. \ref{s:po_dash}).
This is utilized in the following procedure for selecting
consistent points from the equivalence sets \eqref{eqn:equival}.

Three noncollinear object points with corresponding dash representations are selected as reference $R$. Then a consistent choice 
for all other equivalence classes 
can be made by selecting the equivalent with the smallest sum of angle errors to all reference points \figref[f]{fig:steps}:
\begin{align}
    p^O_{ij} &= \arg\min_{p\in P^O_{ij}} \;\; \sum_{(p_r,p'_r)\in R} \;
    \big|\sphericalangle (p, p_r) - \sphericalangle (p^{O'}_{ij}, p'_r)\big|
    \label{eqn:argmin}
\end{align}


For continuous rotational objects such as bottles, a point in the star representation maps to an infinite equivalence class $P^O_{ij}$. 
Methodically, we thus want an infinite $\arg\min$ in \eqref{eqn:argmin}. For practical reasons, this is replaced by
$\tilde{P}^O_{ij}$, which contains for every reference point 
the two possible object points with the desired angle 
$\sphericalangle (p^{O'}_{ij}, p'_r)$.
These points are obtained by first rotating an arbitrary point $\bar{p}^O$ from the
equivalence class above each reference point ($\bar{\bar{p}}^O$). 
These points are then rotated by the angles $\pm\beta$ obtained
by the spherical Pythagorean theorem to get the
desired two points.
\begin{align}
    \tilde{P}^O_{ij} &= \Bigl\{\Rot_Z(\pm \beta)\bar{\bar{p}}^O \Big| (p_r,p'_r)\in R\Bigr\}, 
    \text{ with a } \bar{p}^{O}\in P^O_{ij},
    \label{eqn:pocontinuous}
    \\
     \beta &= \arccos\left(\frac{\cos \sphericalangle (p^{O'}_{ij}, p'_r)}{\cos \sphericalangle (\bar{\bar{p}}^{O}, p_r)}\right)\!\!,\;
        \bar{\bar{p}}^{O} = \cart\mvs{\cyl(p_r)_\phi \\ \cyl(\bar{p}^{O})_\rho \\ \cyl(\bar{p}^{O})_Z}\!\!.
\end{align}
For clarity, this assumes $Z$ as symmmetry axis.

As reference $R$, any three noncollinear object points with corresponding dash representation can be selected, \eg one of the possible object point combinations with the smallest angle error sum for three arbitrary selected output pixel. The rotational axis inside the dash representation can be regressed\footnote[3]{The coordinate along the rotational axis is unchanged in the star representation and therefore available.}. For continuous  rotational  objects, this can be used to form a reference based on the coordinate system, since the other two axis may be selected arbitrarily (if they form a coordinate system).

\section{Experimental 6-DOF Evaluation}
\label{sec:exp_6dof}
We evaluate our approach on the T-LESS Dataset \cite{hodan2017tless} which spotlights 30 industry-relevant objects without discriminative color and texture. Regarding the symmetry the objects can be categorized in eleven $\infty$-fold, $15$ $2$-fold, three $1$-fold and one $4$-fold symmetry around one axis. We accessed the dataset via the "BOP: Benchmark for 6D Object Pose Estimation" which provides standardized simulated training data, evaluation methods and the results from other state-of-the-art algorithms for direct comparison (cf. \cite{hodan2020bop}). Since we only improve the pose estimation, we use the mask R-CNN detector results from \cite{labbe2020cosypose} for evaluation.

\subsection{Network Structure and Learning Procedure}
As network structure we use a DenseNet\cite{huang2017densely}-like encoder-decoder structure with horizontal connections. All (non-output) convolutions are activated by SELU \cite{klambauer2017self}. As optimizer, Adam \cite{kingma2014adam} is used with the amsgrad expansion \cite{reddi2019convergence} and a learning rate of $0.0001$. Our network is trained in two phases: We pretrain the object point relevant outputs for two epochs. Afterwards, we include also the uncertainty outputs. The therefore complete network is then trained for additional ten epochs.
More details can be seen in our implementation.

For training, we generated ten samples for each training datum provided by \cite{hodan2020bop}. For each sample a scale and translation offset is drawn from Gaussian distributions. Additionally, all input images are augmented by contrast, Gaussian and brightness noise and always processed as grayscale images since the objects are colorless.

\subsection{Results}
\begin{table}
\center
\vspace{5px}
\begin{tabular}{lccc}
    Method (RGB) & refinement& \textbf{AR} & AR$_{MSPD}$\\ \hline
    \textcolor{gray}{CosyPose \cite{labbe2020cosypose}}& \textcolor{gray}{RGB} & \textcolor{gray}{72.8} & \textcolor{gray}{82.1} \\
    \textbf{Ours}  & - & 55.2 & 76.5\\
    CDPN \cite{li2019cdpn} & - & 49.0 & 67.4\\
    CDPNv2 \cite{li2019cdpn} & - & 47.8 & 62.0\\
    EPOS \cite{hodan2020epos} & - & 47.6 & 63.5\\
    leaping from 2D to 6D\cite{liu2020Leaping} & - & 40.3 & 71.2\\
    Pix2Pose \cite{park2019pix2pose} & - & 34.4 & 47.6
\end{tabular}
\caption{Average recall (AR) on the T-Less dataset (RGB only, cf. \cite{hodan2020bop})}
\label{tab:6d_results_rgb}

\begin{tabular}{lccr}
    Method (RGB-D) & refinement & \textbf{AR} & \textcolor{darkgray}{ time (s)} \\ \hline
    CosyPose \cite{labbe2020cosypose}  & RGB+ICP & 70.1 &\textcolor{darkgray}{ 13.74}\\
    K\"onig \cite{koenig2020hybrid}  & ICP & 65.5 & \textcolor{darkgray}{ 0.63}\\
    \textbf{Ours} w. depth fusion & -  & 65.1 & \textcolor{darkgray}{ 0.45}\\
    Vidal \cite{vidal2018method}  & ICP & 53.8 & \textcolor{darkgray}{ 3.22}\\
    Pix2Pose \cite{park2019pix2pose} & ICP & 51.2 & \textcolor{darkgray}{ 4.84}\\
    Drost-Edges \cite{drost2010model} & ICP & 50.0 & \textcolor{darkgray}{ 87.57}\\
    Sundermeyer \cite{sundermeyer2020augmented}  & ICP & 48.7 & \textcolor{darkgray}{ 0.86}\\ 
    CDPNv2 \cite{li2019cdpn} & ICP & 46.4  & \textcolor{darkgray}{ 1.46}\\
\end{tabular}
\caption{Average recall (AR) on the T-Less dataset (RGB-D, cf. \cite{hodan2020bop})}
\label{tab:6d_results_rgbd}
\end{table}

Table \ref{tab:6d_results_rgb} shows our average recall (AR, as defined in \cite{hodan2020bop}) on the T-LESS dataset in comparison to other state-of-the-art methods for RGB-only processing.  Our approach reaches state-of-the-art results and is only exceeded by a approache with refinement steps \ie CosyPose \cite{labbe2020cosypose}. Since the T-LESS dataset comprises mainly symmetric objects (28/30), it stands to reason that the proposed approach aids CNNs to converge better.

Since we build upon \cite{richter2019towards}, which introduced a simple method for utilizing the depth image's information by fusing it directly into the PnP stage, we are able to integrate depth data as well. Our results with depth fusion in comparison to state-of-the-art results on RGB-D can be seen in Table \ref{tab:6d_results_rgbd}. We are the only algorithm not refining with an ICP-variant. Therefore, our predictions are calculated noticeably faster (cf. \ref{tab:6d_results_rgbd}). Nevertheless, our results on RGB-D data are competitive.

\section{Related Work}
\label{sec:rel_work}

The problem of symmetry in CNN-based 6D-Pose detection is also discussed in \cite{pitteri2019object}. This work, as well as \cite{rad2017bb8} propose a simple normalization of the pose's rotation. Naturally, this introduces an uncontinuity after one rotation of symmetry, wherefore they furthermore propose to learn a second, offsetted, normalization per symmetry. This normalization is of course also uncontinuous but at a different angle. Finally, a special segmentation is learned in addition to the normalized rotations, which only use is to indicate in which normalization's sweet spot the perceived rotation lies and therefore which normalization output should be used to calculate the pose. This approach is also used in \eg \cite{labbe2020cosypose} or \cite{oberweger2018making}.

Instead of learning 3D object coordinates in one way or another, Hodan \etal~\cite{hodan2020epos} split at first the objects into surface fragments for which then coordinates and probabilities are learned. The probability of one fragment indicates how likely this fragment is seen, given the originating object is observed. Afterwards, the position for each fragment can be calculated and the pose can be extracted by solving a PnP variant over these fragment. Note  that multiple fragments can live next to each other on the same spot, which is only disentangled inside the PnP-RANSAC for many-to-many 2D-3D correspondences. This approach can handle symmetry by learning multiple fragments with the same appearance, which should get  the same probability assigned by the CNN\footnote[4]{Premise: the training data is equally distributed over all symmetries.}. In this approach, the learned coordinates (of the segments) are not biased by uncontinuity as long as the segments are selected sufficiently small since each segment for itself is not symmetric. While this representation inflates the output space, it has the additional advantage of working without knowledge of the object's symmetry. Interestingly, this approach (which is strongly different but also not biased by uncontinuity) reaches highly comparable results to this work (cf. table \ref{tab:6d_results_rgb}). 

The importance of continuity of the rotational representation for a CNN in general was also investigated and affirmed by \cite{zhou2019continuity}, however they did not consider symmetries. 

Peretroukhin et al.\@~\cite{peretroukhin2020smooth} represent rotations implicitly as a quaternion defined by $q^*=\arg\min_{|q|=1} q^TAq$ for a $4\times4$ matrix $A$ which is the output of the network. It defines a Bingham distribution and according to the authors measures uncertainty,
even if instead of likelihood only a loss on $q*$ was
trained. This is related to the $T^*=arg\min_{T\in{}SO(3)}\bar{T}^T(M^TM)\bar{T}$  representation
we use~\cite{richter2019towards} for a rotation matrix $T$ flattened
as $\bar{T}$. Unlike \cite{peretroukhin2020smooth}, it 
represents pose distributions resulting from perspective 
observations.

\section{Conclusions}
\label{sec:concl}

In this work we analysed the effect of symmetric objects on CNN-based pose estimation. We show that without special care, a CNN has to approximate an uncontinuous function which is not optimal. In contrast, we propose a method to warp the CNN's output space in such a way that the uncontinuity is moved to postprocessing outside the CNN. 
Our updated methode reaches state-of-the-art on the T-LESS dataset for unrefining RGB-based methods with an AR of $55.2$.

 \addtolength{\textheight}{-10cm}   


\bibliographystyle{IEEEtran}
{\small
\bibliography{egbib}

\begin{thebibliography}{10}
\providecommand{\url}[1]{#1}
\csname url@rmstyle\endcsname
\providecommand{\newblock}{\relax}
\providecommand{\bibinfo}[2]{#2}
\providecommand\BIBentrySTDinterwordspacing{\spaceskip=0pt\relax}
\providecommand\BIBentryALTinterwordstretchfactor{4}
\providecommand\BIBentryALTinterwordspacing{\spaceskip=\fontdimen2\font plus
\BIBentryALTinterwordstretchfactor\fontdimen3\font minus
  \fontdimen4\font\relax}
\providecommand\BIBforeignlanguage[2]{{%
\expandafter\ifx\csname l@#1\endcsname\relax
\typeout{** WARNING: IEEEtran.bst: No hyphenation pattern has been}%
\typeout{** loaded for the language `#1'. Using the pattern for}%
\typeout{** the default language instead.}%
\else
\language=\csname l@#1\endcsname
\fi
#2}}

\bibitem{richter2019towards}
J.~Richter-Klug and U.~Frese, ``Towards meaningful uncertainty information for
  cnn based 6d pose estimates,'' in \emph{International Conference on Computer
  Vision Systems}.\hskip 1em plus 0.5em minus 0.4em\relax Springer, 2019, pp.
  408--422.

\bibitem{hodan2020bop}
T.~Hodan, M.~Sundermeyer, B.~Drost, Y.~Labbe, E.~Brachmann, F.~Michel,
  C.~Rother, and J.~Matas, ``{BOP} challenge 2020 on 6d object localization,''
  \emph{arXiv preprint arXiv:2009.07378}, 2020.

\bibitem{park2019pix2pose}
K.~Park, T.~Patten, and M.~Vincze, ``Pix2pose: Pixel-wise coordinate regression
  of objects for 6d pose estimation,'' in \emph{Proceedings of the IEEE
  International Conference on Computer Vision}, 2019, pp. 7668--7677.

\bibitem{wang2019normalized}
H.~Wang, S.~Sridhar, J.~Huang, J.~Valentin, S.~Song, and L.~J. Guibas,
  ``Normalized object coordinate space for category-level 6d object pose and
  size estimation,'' in \emph{Proceedings of the IEEE Conference on Computer
  Vision and Pattern Recognition}, 2019, pp. 2642--2651.

\bibitem{wang2019densefusion}
C.~Wang, D.~Xu, Y.~Zhu, R.~Mart{\'\i}n-Mart{\'\i}n, C.~Lu, L.~Fei-Fei, and
  S.~Savarese, ``Densefusion: 6d object pose estimation by iterative dense
  fusion,'' in \emph{Proceedings of the IEEE Conference on Computer Vision and
  Pattern Recognition}, 2019, pp. 3343--3352.

\bibitem{xiang2017posecnn}
Y.~Xiang, T.~Schmidt, V.~Narayanan, and D.~Fox, ``Posecnn: A convolutional
  neural network for 6d object pose estimation in cluttered scenes,''
  \emph{arXiv preprint arXiv:1711.00199}, 2017.

\bibitem{pavlasek2020parts}
J.~Pavlasek, S.~Lewis, K.~Desingh, and O.~C. Jenkins, ``Parts-based articulated
  object localization in clutter using belief propagation,'' \emph{arXiv
  preprint arXiv:2008.02881}, 2020.

\bibitem{hodan2017tless}
T.~Hoda{\v{n}}, P.~Haluza, {\v{S}}.~Obdr{\v{z}}{\'a}lek, J.~Matas, M.~Lourakis,
  and X.~Zabulis, ``{T-LESS}: An {RGB-D} dataset for {6D} pose estimation of
  texture-less objects,'' \emph{IEEE Winter Conference on Applications of
  Computer Vision (WACV)}, 2017.

\bibitem{labbe2020cosypose}
Y.~Labb{\'e}, J.~Carpentier, M.~Aubry, and J.~Sivic, ``Cosypose: Consistent
  multi-view multi-object 6d pose estimation,'' \emph{arXiv preprint
  arXiv:2008.08465}, 2020.

\bibitem{huang2017densely}
G.~Huang, Z.~Liu, L.~Van Der~Maaten, and K.~Q. Weinberger, ``Densely connected
  convolutional networks,'' in \emph{Proceedings of the IEEE conference on
  computer vision and pattern recognition}, 2017, pp. 4700--4708.

\bibitem{klambauer2017self}
G.~Klambauer, T.~Unterthiner, A.~Mayr, and S.~Hochreiter, ``Self-normalizing
  neural networks,'' in \emph{Advances in neural information processing
  systems}, 2017, pp. 971--980.

\bibitem{kingma2014adam}
D.~P. Kingma and J.~Ba, ``Adam: A method for stochastic optimization,''
  \emph{arXiv preprint arXiv:1412.6980}, 2014.

\bibitem{reddi2019convergence}
S.~J. Reddi, S.~Kale, and S.~Kumar, ``On the convergence of adam and beyond,''
  \emph{arXiv preprint arXiv:1904.09237}, 2019.

\bibitem{li2019cdpn}
Z.~Li, G.~Wang, and X.~Ji, ``Cdpn: Coordinates-based disentangled pose network
  for real-time rgb-based 6-dof object pose estimation,'' in \emph{Proceedings
  of the IEEE International Conference on Computer Vision}, 2019, pp.
  7678--7687.

\bibitem{hodan2020epos}
T.~Hodan, D.~Barath, and J.~Matas, ``Epos: Estimating 6d pose of objects with
  symmetries,'' in \emph{Proceedings of the IEEE/CVF Conference on Computer
  Vision and Pattern Recognition}, 2020, pp. 11\,703--11\,712.

\bibitem{liu2020Leaping}
J.~Liu, Z.~Zou, X.~Ye, X.~Tan, E.~Ding, F.~Xu, and X.~Yu, ``Leaping from 2d
  detection to efficient 6dof object pose estimation.'' \emph{ECCVW}, 2020.

\bibitem{koenig2020hybrid}
R.~Koenig and B.~Drost, ``A hybrid approach for 6dof pose estimation.''
  \emph{ECCVW}, 2020.

\bibitem{vidal2018method}
J.~Vidal, C.-Y. Lin, X.~Llad{\'o}, and R.~Mart{\'\i}, ``A method for 6d pose
  estimation of free-form rigid objects using point pair features on range
  data,'' \emph{Sensors}, vol.~18, no.~8, p. 2678, 2018.

\bibitem{drost2010model}
B.~Drost, M.~Ulrich, N.~Navab, and S.~Ilic, ``Model globally, match locally:
  Efficient and robust 3d object recognition,'' in \emph{2010 IEEE computer
  society conference on computer vision and pattern recognition}.\hskip 1em
  plus 0.5em minus 0.4em\relax Ieee, 2010, pp. 998--1005.

\bibitem{sundermeyer2020augmented}
M.~Sundermeyer, Z.-C. Marton, M.~Durner, and R.~Triebel, ``Augmented
  autoencoders: Implicit 3d orientation learning for 6d object detection,''
  \emph{International Journal of Computer Vision}, vol. 128, no.~3, pp.
  714--729, 2020.

\bibitem{pitteri2019object}
G.~Pitteri, M.~Ramamonjisoa, S.~Ilic, and V.~Lepetit, ``On object symmetries
  and 6d pose estimation from images,'' in \emph{2019 International Conference
  on 3D Vision (3DV)}.\hskip 1em plus 0.5em minus 0.4em\relax IEEE, 2019, pp.
  614--622.

\bibitem{rad2017bb8}
M.~Rad and V.~Lepetit, ``Bb8: A scalable, accurate, robust to partial occlusion
  method for predicting the 3d poses of challenging objects without using
  depth,'' in \emph{Proceedings of the IEEE International Conference on
  Computer Vision}, 2017, pp. 3828--3836.

\bibitem{oberweger2018making}
M.~Oberweger, M.~Rad, and V.~Lepetit, ``Making deep heatmaps robust to partial
  occlusions for 3d object pose estimation,'' in \emph{Proceedings of the
  European Conference on Computer Vision (ECCV)}, 2018, pp. 119--134.

\bibitem{zhou2019continuity}
Y.~Zhou, C.~Barnes, J.~Lu, J.~Yang, and H.~Li, ``On the continuity of rotation
  representations in neural networks,'' in \emph{Proceedings of the IEEE
  Conference on Computer Vision and Pattern Recognition}, 2019, pp. 5745--5753.

\bibitem{peretroukhin2020smooth}
V.~Peretroukhin, M.~Giamou, D.~M. Rosen, W.~N. Greene, N.~Roy, and J.~Kelly,
  ``A smooth representation of belief over so (3) for deep rotation learning
  with uncertainty,'' \emph{arXiv preprint arXiv:2006.01031}, 2020.

\end{thebibliography}
}

\end{document}